\documentclass[authorversion,nonacm]{acmart}
\AtBeginDocument{%
  \providecommand\BibTeX{{%
    \normalfont B\kern-0.5em{\scshape i\kern-0.25em b}\kern-0.8em\TeX}}}

\setcopyright{acmlicensed}
\copyrightyear{2018}
\acmYear{2018}
\acmDOI{XXXXXXX.XXXXXXX}

\acmConference[FAccT `24]{2024 ACM Conference on Fairness, Accountability, and Transparency (ACM FAccT)}{June 3-6, 2024}{Rio de Janeiro, Brazil}
\acmISBN{978-1-4503-XXXX-X/18/06}

\begin{document}

\title{Discipline and Label: A WEIRD Genealogy and Social Theory of Data Annotation}
\author{Andrew Smart}

    \affiliation{%
    \institution{Google Research}
    \country{US}}
  \email{andrewsmart@google.com}
  
  \author{Ding Wang}
      \affiliation{%
    \institution{Google Research}
    \country{US}}

  \author{Ellis Monk}
     \affiliation{%
    \institution{Harvard University \& Google Research}
        \country{US}}

  \author{Mark Díaz}
 \affiliation{%
    \institution{Google Research}
        \country{US}}

  \author{Atoosa Kasirzadeh}
 \affiliation{%
    \institution{University of Edinburgh \& Google Research}
        \country{US}}

  \author{Erin van Liemt}
 \affiliation{%
    \institution{Google Research}
        \country{US}}

 \author{Sonja Schmer-Galunder}
 \affiliation{%
    \institution{University of Florida}
        \country{US}}

\renewcommand{\shortauthors}{Smart, et al.}

\begin{abstract}
  Data annotation remains the \emph{sine qua non }\footnote{Without which not: an essential condition; a thing that is absolutely necessary.} of machine learning and AI. Recent empirical work on data annotation has begun to highlight the importance of rater diversity for fairness, model performance, and new lines of research have begun to examine the working conditions for data annotation workers, the impacts and role of annotator subjectivity on labels, and the potential psychological harms from aspects of annotation work. Data annotation has become a global industry. This paper outlines a critical genealogy of data annotation; starting with its psychological and perceptual aspects - what exactly are data annotators doing? What are data annotations of? We draw on similarities with critiques of the rise of computerized lab-based psychological experiments in the 1970's which question whether these experiments permit the generalization of results beyond the laboratory settings within which these results are typically obtained. These computerized tests enabled standardized presentation of stimuli and measures of accuracy and response times at the expense of ecological validity. Similarly, do data annotations permit the generalization of results beyond the settings, or locations, in which they were obtained? Moreover, Western psychology is overly reliant on on participants from Western, Educated, Industrialized, Rich, and Democratic societies (WEIRD). Many of the people who work as data annotation platform workers, however, are not from WEIRD countries; most data annotation workers are based in Global South countries. Social categorizations and classifications from WEIRD countries are imposed on non-WEIRD annotators through instructions and tasks, and through them, on data, which is then used to train or evaluate AI models in WEIRD countries. Thus, another question is; what does it mean for non-WEIRD workers to annotate data from and about WEIRD societies? Is there an inverse WEIRD effect? We synthesize evidence from several recent lines of research and argue that data annotation is a form of automated social categorization that risks entrenching outdated and static social categories that are in reality dynamic and changing. We propose a framework for understanding the interplay of the global social conditions of data annotation with the subjective phenomenological experience of data annotation work.      
\end{abstract}

\maketitle

\section{Introduction}

Despite the swift and significant progress in artificial intelligence, it remains an inescapable fact that these AI systems fundamentally rely on humans for labeling the data essential for their training, evaluation, fine tuning, and adversarial testing. Collecting large amounts of labeled data remains an indispensable part of the machine learning pipeline, and progress in AI has been driven by labeled data and compute. There is a pervasive view that optimizing for inter-rater reliability on large amounts of data overrides the qualitative aspects of the annotation process. Data has been, according to some, unreasonably effective \cite{halevy2009unreasonable}. Unreasonably effective for \textit{what} is less clear.  Nevertheless, without countless human annotations, the meteoric rise of AI could not have happened and effective machine learning results would be impossible. 

\subsection{Elusive ground truth}
\begin{quote}
    "Any assertion regarding facts, even the simplest...is already an interpretation...and therefore itself already a theory" - Hans Reichenbach 
\end{quote}

The "ground truth," the foundation of machine learning models, is essentially a construct of human consensus \cite{denton2021whose}. In practice, this "truth" is often a reflection of the perspectives and interpretations of a specific cohort of domain experts and model developers, who then convey these definitions to another group, typically the data annotation workers. This process highlights the subjective nature inherent in the establishment of ground truths within AI systems. Crucially, the ground truth for machine learning systems is a socially and human-constructed concept \cite{jaton2021constitution}. As others in the FAccT community have argued, ground truth in annotation is not universal and disagreement among annotators can provide a lens into social and cultural differences \cite{davani2022dealing, basile2021toward}. However, using genealogy as method we deepen this critique and orient our critique toward the fundamental nature of the project of data annotation, which serves as a technologically-mediated form of social categorization, regardless of how it is implemented or how resulting annotations are used. The ground truth in computer science does not exist independently of the minds of algorithm developers and human data annotation workers. The question is who gets to decide what ground truth AI systems are learning \cite{diaz2022crowdworksheets}?

As we will unpack, machine learning is presumed to be both a truth-seeking and a knowledge-generating enterprise \cite{smart2020reliabilism}. It's worth noting that the epistemic standards used to evaluate scientific knowledge differ from those applied to AI systems. In fact, these standards are generally lower when ascribing knowledge to machines\cite{boge2022two}. The purpose of machine learning is about what can be inferred from data. However, machine learning seeks to circumvent the requirement to explicitly set modeling assumptions prior to drawing inferences from a model's output \cite{wheeler2016machine}. For the sake of technical simplicity, machine learning operates under the assumption that there is a single, ostensibly neutral ground truth for the concepts being annotated \cite{kapania2023hunt, aroyo2023dices}. 

\subsection{The value-laden nature of data annotation and machine learning}
A persistent and common folk misconception among AI practitioners is that machine learning involves objective math and data and is therefore free of the interference of cultural values, personal speculation or emotional interest \cite{johnson2023algorithms, zeerak2020disembodied}. However, the last decade of research has demonstrated that machine learning algorithms are in fact heavily laden with the values of their creators \cite{birhane2022values, andrews2023devil}. As Cathy O'Neil said, “Models are opinions reflected in mathematics” \cite{o2017weapons}. These values also motivate how datasets are created, how problems to work on are chosen, what counts as data, and what counts as "quality" data \cite{paullada2021data}. When those datasets have to be labeled, the values of the requesters of the labels are reflected in the task instructions given to the data annotation workers \cite{miceli2020between,wang2022whose, vij2023women}. Ultimately, the determination of "correct" labels is often in the hands of a limited group of individuals within large technology companies and leading institutions.    

Having established that social and cultural	values	play	an	important	role	in	
machine learning,	the	key	question	for	anyone	who	advocates	this	view	is:	\emph{what}	and \emph{whose} values \cite{psillos2015evidence}?
If the values imposed on annotators are from Western, Educated, Industrialized, Rich, Democratic (WEIRD) societies \cite{henrich2020weirdest}, the "ground truth" against which model performance is benchmarked would equal WEIRD values. The WEIRD problem in psychology was identified a decade ago and refers to the fact that the field is overly reliant on participants from Western, Educated, Industrialized, Rich, and Democratic societies \cite{de2023psychology}. But it turns out that the lack of diversity in psychology is a symptom of much deeper epistemic problems, to which we will return. 

And indeed there is ample evidence for example that the latest generation of Large Language Models and Text2Image models reflect Western values \cite{benkler2023assessing, johnson2022ghost, qadri2023ai}. Social categorizations and classifications from WEIRD countries are imposed on non-WEIRD annotators, and through them, on data, which is then used to train, fine tune or evaluate AI models in WEIRD countries \cite{miceli2020between, diaz2022crowdworksheets, wang2022whose}. Critical data studies has also examined how data
is never ‘raw’ \cite{sambasivan2021everyone, miceli2020between, gitelman2013raw}, but is shaped through the practices of
collecting, annotating, curating and sensemaking, and thus is inherently sociopolitical in nature. Indeed, raw unlabeled or uncleaned data is neither economically nor mathematically valuable at all; this is why the entire data annotation industry exists.

These models are then used to automate decisions or make predictions on people in WEIRD countries (and beyond), which, as is well-documented, can lead to myriad social harms, especially for already structurally vulnerable people \cite{shelby2023sociotechnical, maalsen2023algorithmic}. This remarkably complex feedback loop has been under-theorized, and one aim of this paper is to unpack aspects of it building on the growing body of scholarship around data annotation work \cite{diaz2022crowdworksheets, wang2022whose, denton2021whose}.

The values inherent in data annotation practices undergo a complex interplay of interpretation across languages and cultures, particularly when this task is outsourced to countries with lower labor costs through technology platforms and business process outsourcing (BPO) companies. However, the annotators who are directed to apply these assigned labels are seldom given the opportunity or channel to raise questions or challenge them. This situation perpetuates the prevailing notion that the financial patron of the project possesses the authority to dictate the assigned meanings \cite{miceli2020between}.  

\subsection{AI is non-WEIRD people}
The automated and intelligent products that we use in our daily lives, developed and marketed by companies in the Global North, are fundamentally dependent on the efforts of millions of data annotation workers, predominantly from the Global Souths\cite{vij2023women}. In contrast to the celebrity status of AI researchers, the data annotation workers whose aggregated judgements enable AI to work remain largely anonymous \cite{miceli2020between, gray2019ghost, vij2023women}. We also theorize what exactly the act of data annotation is---a form of technology-mediated social categorization.



There is increasing focus on the labor practices and examining the power structures in the data industry in the algorithmic fairness, CHI and FAccT literature \cite{miceli2020between, miceli2022data, diaz2022crowdworksheets, paullada2021data, kapania2023hunt}. Through ethnographic methods and social theory these works have revealed data annotation to often be precarious while at same time data workers can be under tight surveillance and control \cite{gray2019ghost, wang2022whose}. Data annotation work is seen as unskilled, yet high-quality and accurate annotation is expensive, time-consuming, and fraught with exploitative, unregulated labor practices in the very countries where data is scarcest\cite{gray2019ghost}. Additionally, integrating and acknowledging the lived experiences and identities of data annotators and understanding how these factors influence their judgments present a significant challenge in the field of machine learning. \cite{diaz2022crowdworksheets, wang2022whose}. 

\subsection{Contributions}
Our fundamental question we seek to investigate is: what are the implications of having people from very different societies, located in disparate nodes in the global economy, make social judgements about people in other societies mediated by algorithmic systems? How does the economic and social power of the data requesters influence the work of data annotation? We argue that a holistic view of data annotation must take into account the way in which social identities are manifest - and this requires very specific context-based analysis \cite{alcoff2005visible}. As Alcoff points out, identities are constituted by social contextual conditions of interactions in specific cultures at particular historical periods \cite{alcoff2005visible}. However, algorithmically mediated social judgements about people's identities in different societies necessarily make generalized claims about social categories as the entire purpose of machine learning is generalization.   

In light of the above arguments, and with a concern for epistemic and social justice in AI, we make the following contributions: 
\begin{itemize} 

    \item We propose a comprehensive theoretical framework synthesizing the expanding body of literature on data annotation. This framework operates on these two distinct but interconnected levels. First at the micro-level analysis, individual psychology and perception, we examine the cognitive and perceptual aspects of data annotators. Second, at the macro-level analysis, social structures and global dynamics, we explore the broader societal and economic forces that shape the data annotation industry. 
    \item We draw parallels and critical comparisons with statistical inference in WEIRD experimental psychology and the rise of machine learning.
    \item We propose an account for how to understand the interplay of social categories, social positions, and machine learning.  
\end{itemize}

\section{what exactly is data annotation?}
Data annotation workers carry out a range of tasks fundamental to AI and machine learning; from data labeling to text and image transcribing \cite{le2023problem, vij2023women}. Moreover, data work, of which labeling is a part, receives relatively little attention compared to other stages of model development \cite{sambasivan2021everyone}. As we explain below, much of these tasks are essentially forms of social categorization - a set of processes whereby human beings make judgments and inferences about how various entities belong to myriad categories or adequately represent certain concepts \cite{mccrae2005universal}. Data annotation tasks are commonly performed by workers from low-income countries, who often earn poverty wages and the work is part of a large shift in the global economy toward precarious piecework \cite{le2023problem, gray2019ghost}. Reliable estimates are hard to come by as statistics on data annotators are often proprietary.  Crowdsourced data annotation work on large web-based platforms is additionally following the gig-work model where there are few permanent jobs, fewer healthcare or retirement benefits, and typically a lack of possibilities for union organization or establishing worker protection \cite{gray2019ghost}.

Data annotation is sometimes referred to as labeling or rating, and the people who perform this work are called “annotators”, “labellers” or “raters”, but also "crowdworkers", "microworkers", "microtaskers", "gig workers" or "online freelancers" \cite{vij2023women}. These terms are overlapping and largely interchangeable and at bottom these terms refer to the act of humans performing categorical judgements on image, video, text, audio or other data in order to provide machine learning algorithms with the above mentioned “ground truth” against which to measure predictive accuracy. We follow Asmita Vij \cite{vij2023women} in calling for the demystification of data annotation work: the producer of a commodity (labelled data) and the means to used to produce that commodity (platforms), an accurate phrase to describe these workers is data annotation platform workers. 

Data annotation has become part of the infrastructure of AI \cite{sambasivan2021everyone, crawford2018anatomy}, and like physical infrastructure, few people pay attention to the politics and power relationships that these infrastructures embody.  Yet, the data labelling industry as infrastructure is part of the background, necessary for the functioning of the tech industry and AI research but rarely noticed or discussed. Much in the same way that the global supply chain for AI chips is a crucial part of enabling the AI “revolution”, so is data labelling, but its functioning and conditions of possibility are rarely noted or studied by the very field whose entire existence is utterly dependent upon it. 

The spread of data annotation has a dual nature of increasing global market dependence for dispossessed people, in that it produces and maintains the imperative to work low-wage jobs to meet basic needs \cite{franklin2021digitally}. It is a form of social automation. Due to the global digital reach of platforms that connect labor markets (technologists who need their data annotated at the lowest possible price) with workers (who need their basic needs met), data annotation redistributes work to its smallest reducible parts and lowest possible bids \cite{roberts2019behind}. This kind of app-driven gig labor has deeply racialized and gender-based characteristics of exploitation \cite{vij2023women, roberts2019behind}. 

This \emph{capital relation}, which is the relationship between the data annotation platform worker who sells their labor (in this case their cognitive labor) and the technology firms who buy it, appears at first glance entirely voluntary \cite{mau2023mute}. And in mainstream machine learning research and economics the relations between data annotation platform workers and the platforms is treated as such - a voluntary market exchange of cognitive labor. However, a simple glance at the conditions under which this relationship exists reveals that it is in fact a relationship of domination and exploitation \cite{vij2023women, mau2023mute, miceli2020between}.


\subsection{Genealogy as a method: examining power and making labor visible}
Genealogy as a philosophical method seeks to see through the rationalized surface of traditional examinations of social and technological phenomena like data annotation, to the actual human beings who lie behind it \cite{solomon2012nietzsche}. We argue that one can only truly understand a phenomenon when we understand its origins, its development, and its overall place in larger social structures and forces \cite{haslanger2012resisting}. Such an understanding is necessary to capture how transnational platforms owned by large tech firms are impacting the economic, social, and political lives of workers across the globe \cite{vij2023women}. This perspective is also necessary to understand how the automated social categorization algorithms impact the lives of structurally vulnerable people on the other end of the machine learning predictions: workers, prisoners, patients, students, teachers, drivers, and anyone else who is structurally dependent on the output of algorithms to meet their daily needs \cite{o2017weapons}. Our aim, however, is not to provide a single, unified history of data annotation, but instead to point to the complex epistemological issues, economic processes, professional accidents and contingencies that underlie data annotation \cite{cryle2019normality}. We wish to open new lines of inquiry about data annotation that go beyond narrow technical and bias concerns, building on power-aware research \cite{miceli2022studying}.  

We also see the ethical and political obligations of the FAccT community to extend beyond those who belong to our own local, regional, national or intellectual community, or to our own cultural group, to include data annotation platform workers - according to a \emph{social connection model of responsibility} defended by Iris Young and Jose Medina \cite{young2010responsibility, medina2012epistemology}. This point is especially salient as FAccT itself is overly reliant on perspectives from WEIRD societies \cite{septiandri2023weird}.

One of our goals with this paper is to drag many unexamined parts of data annotation as an act into the light and scrutinize them; to make the hidden aspects of data annotation, such as epistemic values, worker positionalities, visible rather than an unquestioned part of the machine learning supply chain.  This builds on recent work such as Data Feminism which calls on data science researchers to examine and challenge power as well make the labor behind data work visible \cite{d2023data}, and Miceli et al's introduction of a power-oriented perspective that highlights the dynamics of imposition and naturalization inscribed in the classification, sorting,
and labeling of data \cite{miceli2020between}. As with previous work on the genealogy of machine learning datasets \cite{denton2021genealogy}, and work critically examining dataset construction \cite{paullada2021data}, the goal of this genealogical analysis of data annotation is to deepen the practice of critical self-reflection about under-examined aspects of machine learning.

Prior genealogical analysis and critical history of ImageNet, for example, focused on excavating assumptions around the aggregation and accumulation of more data, the computational construction of meaning, and making certain types of data labor invisible \cite{denton2021genealogy, crawford2018anatomy}. Here we focus the latter. Building off of this work, our novel contribution is to place the existing empirical findings about data annotation workers into a theoretical framework that draws on the history of psychology and the sociology of social categorization. 

\subsection{The Objectivity of the Subjective in Data Annotation}
Subjectivity exists as part of the objective world, and therefore to grasp the world objectively one must grasp the subjective \cite{reed2011interpretation}. Bourdieu's conception of \emph{habitus} which seeks to understand the co-creation of 'objective' material social reality - e.g., the actual distribution of capital, the existence of economic classes - with 'subjective' phenomenological reality - the perception and recognition of these material realities \cite{bourdieu1990logic}. Bourdieu argues that it is a mistake to examine these in isolation form each other - the objective social world and the subjective experience of this world. \emph{Habitus}, according to Bourdieu, consists of social systems of durable principles which generate and organize social practices, as well as the representations of these practices \cite{bourdieu1990logic}, \emph{habitus} as the background conditions of social life for Bourdieu determines what is thinkable and unthinkable in any given society.   

Data annotation is a salient example where a concept like \emph{habitus} is explanatory, as data annotation is a site of subjectivity and objectivity colliding. As we've mentioned, ML models were initially hailed
as objective, unimpeded by subjective human biases \cite{zeerak2020disembodied}. And ostensibly, the purpose of annotating so-called raw data is to transform it into something more valuable, a commodity \cite{vij2023women}. But data annotation also serves as the basis for training machine learning algorithms, in order that they can learn what it is that humans wish them to do, which is to automate social categorization, classifications, predictions, decisions, and even scientific discovery \cite{boge2022two}. 

What happens though when the classifications and predictions we wish to automate by teaching machine learning algorithms through data annotations are social categories? We wish to automate social categorization. This connects to Hacking's paradigm of socially constructed categories, where people draw from socially available classifications into their intentional agency and sense of self, thereby changing the categories themselves, and thus the classifications evolve with them \cite{hacking2013making, haslanger2012resisting}. Thus, the subjective identities and social positions of the data workers and must play a key role in any account of how machine learning systems come to behave the way do \cite{miceli2020between}. 

The social position of "data annotation worker" entails a broad range of norms, obligations, and expectations, and is embedded in a matrix of practices and institutions \cite{haslanger2012resisting}. For example, the data annotation industry follows the outsourced low-cost globalized "labor arbitrage" model, where labor that is seen as unskilled in high-cost countries is sent to cheap-labor countries, similar to manufacturing and other physical industries \cite{vij2023women, peck2017offshore, friedman2008modernities}. Production, as well as data annotation work, is relocated to areas of cheap labor, lower taxes, less labor regulation, and better financial conditions \cite{friedman2008modernities}. Or it follows the precarious gig-platform worker model which, while more geographically distributed, is nonetheless precarious and usually low-paid \cite{gray2019ghost}. Thus, a concern for justice, ethics, fairness in machine learning, requires an examination of whether and why being socially positioned as a data annotation platform worker in machine learning is a subordinated status.

\section{ Data Annotation and a detour through WEIRD experimental psychology}
In many ways the act of data annotation resembles how experimental psychologists have studied human cognition since the advent of the personal computer. Prior to widespread access to personal computers, psychological experiments were often carried out with pencil and paper, asking human participants to perform repetitive tasks and measuring the mean differences in performance between groups or tasks . For example, one of the most well-known psychological laws is Fitts Law, which measures the speed-accuracy trade-off in fine motor movements - the faster you move the less accurate you are \cite{fitts1954information}. The foundational experiments that led to the formulation of Fitts's Law involved participants performing tasks with a pencil and paper, while an experimenter measured the time taken using a stopwatch. Despite the simplicity of this approach, the data collected from these experiments enabled Fitts to come up with a highly predictive equation that became Fitts Law which remains empirically confirmed today.  

As computers became more powerful, cheaper and more widely available, psychologists began using computer-based experiments to measure human performance on a wide range of tasks like perceptual judgment, memory span or cognitive function \cite{musch2000brief}. This allowed researchers to measure human behavior with unprecedented precision. Although it is at the expense of  “ecological validity”. The artificial setting of a university lab, where participants might click on small red squares on a screen or recall lists of words, hardly mirrors the cognitive demands of everyday activities like conversing with someone, cooking dinner, or navigating a busy street. 

Psychological tasks such as the Stroop Task, N-back, or Flanker test, commonly used in experiments, hardly resemble the kinds of cognitively demanding activities humans routinely perform in their daily lives, like getting dressed, gardening, or collectively doing large-scale software engineering projects. Yet understanding these everyday activities is ostensibly the aim of psychology. These laboratory tasks are based on the assumption that they tap into underlying connive functions related to behaviors like reading, which are of real interest to researchers. For example, the ability to complete the Stroop task is thought to reveal aspects of cognitive processing that are crucial for reading. Some researchers have even studied how it is possible that humans could come into a psychology lab and perform the arbitrary and contrived tasks concocted by psychologists \cite{posner2004attention}. Similarly, we may also wonder how is it that humans can be arbitrarily instructed to attach labels to data examples. Despite this, lab-based computerized psychological experiments have led to highly accurate and precise measurements of psychological constructs like working memory, attention, conscious awareness and long-term memory. This data has in turn enabled computational cognitive theories to become more detailed and rich. 

\subsection{Generalizing from the observed to the unobserved: psychology, AI and the problem of induction}

\begin{quote}
    "But in fact, we know nothing from having seen it; for the truth is hidden in the deep."
    
    - Democritus
\end{quote}
A key assumption of lab-based psychology experiments remains that performance on these tasks would not be influenced by social factors or cultural difference. This assumption underpins the practice of generalizing findings from specific study populations to humanity as a whole. In essence, these experiments are believed to capture the behavior and thought processes common to all humans \cite{de2023psychology}. However, this approach is challenged by a philosophical question, posed by Hume: How can we extrapolate conclusions beyond the specific experiences, observations, or data we have encountered, to broader contexts we have not directly experienced \cite{hume2000treatise, popper2013realism}?

To address the problem of induction, psychology turns to \textit{inferential} statistics, a cornerstone methodology for deriving broader conclusions from specific data sets. Textbooks often extol statistics as ``a method of pursuing truth'' and further assert that "this pursuit of truth, or at least its future likelihood, is the essence of psychology, of science, and of human evolution." \cite{aron1999statistics}. Unlike \textit{descriptive} statistics, which focus on summarizing data through measures like averages, inferential statistics enable psychologists to make broader inferences from their study data. This approach involves drawing conclusions about a larger population based on a sample, a practice that is strikingly similar to what occurs in machine learning. Machine learning algorithms use data annotations from a relatively small group of individuals to infer social categories, applying these insights to a much broader population, potentially spanning diverse societies.

Interestingly, the institutionalization of inductivism and inferential statistics as essential to the scientific method in psychology developed in concurrence with the birth of the field of AI --- around 1940 - 1955 \cite{gigerenzer1989empire}. It was during this time that the idea of creating machines that can "think" or act like humans do, or ought "rationally" to do, began to coalesce into a discipline \cite{wheeler2016machine, dupuy2009origins}. A critical aspect of these parallel intellectual developments is their lack of inclusivity, notable the exclusion of women, people of color, and individuals outside Europe and the United States \cite{devlin2023power}. It has taken decades of feminist standpoint epistemology, critical studies and Black feminist scholarship to to reveal how this ostensibly scientific and universalizing worldview---claiming to define universal truths about humanity---actually reflects the perspectives of a narrow demographic of white male academics \cite{longino1990science, harding2004feminist, benjamin2023race, sutherlin2023human}. 


In essence, both machine learning and psychology grapple with a rather significant problem of induction \cite{psillos2007philosophy}. This inductive inference is question-begging, assuming the very regularities of human behavior and social categories it seeks to demonstrate. If there are universal human regularities then generalization from specific samples is justified, however we cannot empirically demonstrate these universals. Again, the fundamental goal of machine learning is \textit{generalization} from examples \cite{mohri2018foundations}.

\subsection{Things get WEIRD}
The practice of generalizing experimental, psychological results from a narrow participant pool---predominantly WEIRD undergraduate students performing contrived tasks in lab settings---to the real world is already scientifically tenuous. The leap from these Western-centric experimental contexts to the entirety of human diversity is even more precarious \cite{henrich2020weirdest, de2023psychology}. If the participant pool of the vast majority of psychological and behavioral economic studies fails to reflect global human diversity, the reliability and universality of the conclusions drawn in these disciplines become questionable\cite{de2023psychology}. Even though this issue has long been recognized within psychology, it has only recently been given formal attention in the literature \cite{henrich2020weirdest}. Despite this recent attention, it appears that psychologists have not changed much about their experimental practices in terms of recruiting participants from other cultures or populations \cite{de2023psychology}. What does this problem in psychology imply for data annotation? 

\subsection{Big data to the rescue?}
With the advent of the personal computer, the internet, mobile devices, and more powerful computers - enabling crowdwork and large scale data annotation - psychology began to turn from explanatory theories of the mind to focusing on collecting more and bigger data to predict behavior \cite{jones2016big}. In response to the WEIRD problem, which has received a lot of attention in psychology, psychology experienced a swift adoption of online platforms for experimental research\cite{de2023psychology}. However, the surge in data availability introduced a new challenge:the frequent conflation of explanation and prediction in both machine learning and psychology \cite{shmueli2010explain, boge2022two}. While the foundational problem of induction---the philosophical doubt about justifying the belief that observed data will mirror unobserved data, as articulated by Hume---is seldom explicitly acknowledged in psychology and economics \cite{hume2000treatise}, there remains a persistent unease. Many suspect that claims of universalization and generalization from these fields might be unjustified at best, and wrong at worst. This skepticism has driven the desire of ever-larger datasets under the assumption Big Data might rectify issues such as the limitations of small WEIRD study populations \cite{henrich2020weirdest}, sampling bias, and replication crisis in these fields\cite{kosinski2016mining}. Both machine learning and the social and behavioral sciences are heavily influenced by \emph{inductivism}, a philosophy advocated by John Stuart Mill, which posits induction as the fundamental basis of knowledge and asserts its self-justifying nature \cite{psillos2007philosophy}. These theoretical and even metaphysical commitments in psychology are also fundamentally part of what can be termed WEIRD epistemology \cite{de2023psychology}. 


A similar assumption underpins data annotation, in that data annotation workers are seen as interchangeable units of human computation \cite{diaz2022crowdworksheets}. The individual judgements made by human annotators could in principle (on assumption) be made by any human. This assumption has been challenged by recent work showing how individual subjectivity and social identity of the raters influences the annotations \cite{aroyo2023dices, prabhakaran2021releasing}. 

\section{Biased annotators, biased instructions or biased models?}
Dataset annotation work practices and labor conditions have been garnering increasing attention. 
Ever since the ground-breaking gender shades work in which \cite{buolamwini2018gender} showed that facial recognition systems are systematically less accurate for darker-skinned women in particular - performing at near chance, an avalanche of subsequent research has uncovered myriad biases in machine learning systems . Much of this has focused on how subjective values, judgments, and biases of annotators contribute to undesirable or unintended dataset bias \cite{paullada2021data}. The individual subjectivity of data annotators has been proposed as one source of algorithmic bias \cite{miceli2020between}.  The relationship between annotator bias and how biased outputs arise in supposedly atheoretical \cite{andrews2023devil}, value-free \cite{johnson2022ghost}, and purely data-driven algorithmic systems remains an active area of research \cite{geva2019we, al2020identifying, parmar2022don, sap2021annotators}. 

In general, a fundamental assumption in data annotation is that there exists exactly one correct label for every instance of data, and this correct label can be ascertained by as few as three human data annotators. This, however, is just one of the many myths associated with data annotation \cite{aroyo2015truth}. Unfortunately, the belief that there is some "ground truth" out there in the world to be found even holds in cases where such a notion is totally inappropriate (e.g., many instances of semantic interpretation). As we explain below, the implications of this myth become even more pernicious when it pertains annotations that are sensitive to cross-cultural variation.

In fact, not only are these human universals often difficult to empirically confirm, psychologists have now turned to developing techniques to empirically assess cultural distance between various societies \cite{muthukrishna2020beyond}. As psychologists have noted, the discipline's body of research remains dominated by Western, educated, industrialized, rich, and democratic (WEIRD) countries. This raises the question of how generalizable the findings from studies conducted in these countries with samples of study participants almost exclusively from these countries actually is compared to non-WEIRD countries. Furthermore, studies find meaningful cultural differences even among WEIRD countries \cite{mccrae2005universal}. In short, not only is there, then, significant cultural variation in various psychological traits, but also, as it should be obvious, in the social meanings of and social categorizational processes related to race/ethnicity, gender, sexuality, sentiments (e.g., what is or is not perceived as offensive, positive, negative, neutral and how these sentiments are expressed), and more.

Indeed, research shows that ignoring cross-cultural variation has warped LLMs to behave in ways that show profound WEIRD bias \cite{benkler2023assessing}. "Social norms inform us what physical and psychological tools to use to solve recurrent problems depending on the socio-ecological and interpersonal contexts we are embedded in, hence producing substantial psychological diversity around the globe. A consequence of this is that LLMs have inherited a WEIRD psychology in many attitudinal aspects (e.g., values, trust, religion) as well as cognitive domains (e.g., thinking style, self-concept)" \cite{atari2023humans}. These problems are so pervasive that they appear even in multilingual LLMs responding to prompts in non-English languages \cite{atari2023humans}. These issues have led some researchers to "add an amendment to the 'stochastic parrot' analogy and argue that LLMs are a peculiar species of parrots, because their training data are largely from WEIRD populations: an outlier int he spectrum of human psychologies on both global and historical scales. The output of current LLMs on topics like moral values, social issues, and politics would likely sound bizarre and outlandish to billions of people living in less-WEIRD populations" \cite{atari2023humans}.  

\subsection{Inverse WEIRD?}
Ironically, in contrast to LLMs, data annotation essentially produces labels (data) from less-WEIRD, lower income populations (annotators) of entities from mostly-WEIRD countries. This raises the likely possibility that ignoring cross-cultural variation in this context may result in an inverse WEIRD dynamic – annotation largely from the Global South in non-WEIRD countries of entities (e.g., photos, video, text, etc.) largely from the Global North in WEIRD countries. Certainly one implication here is that the assumption that underpins data annotation, which is that data annotation platform workers are interchangeable units of human computation seems untenable. At the very least, it seems problematic to assume that individual judgments made by human annotators could be made by any human, given all of the above and recent work showing how individual subjectivity and the social identity of raters influences their annotations. 

However, this assumption may be violated by for example instruction bias, which are written by the dataset creators or technology organizations which send out their datasets to be labelled \cite{parmar2022don}. 
The precise causal relationship of biases found in the training data - whether it is from under-representation or mis-representation of particular groups - to biased outputs also drives current research. Understanding how social dynamics and structures give rise to data generating processes is a critical new area of sociotechnical research \cite{prabhakaran2020participatory}. Human annotators are one mechanism through which accuracy biases can emerge in machine learning systems \cite{aroyo2023dices, aroyo2015truth, diaz2022crowdworksheets, ghai2020measuring}. 




\section{Data annotation, seeing like an algorithm and social categorization} 
A very typical task that data annotation workers are asked to do is to label data instances for social identities or categories, and this is often done in the service of the goal of fairness \cite{miceli2020between, schumann2021step}. For example, if a model developer wants to know whether AI-generated text or images reproduce unwanted associations between a marginalized identity group and a stereotypical imagery, annotators may be used to make determinations about which identity groups are depicted in a selection of data or to judge whether certain associations are stereotypical or harmful.

As we opened with at the top of the paper, we argue that data annotation is algorithmically mediated social categorization. This requires a brief discussion of theories of social categories and what this means for data annotation platform workers who are tasked with making these kinds of judgements on data about people whom the workers have never met, and are potentially from distant societies - both geographically and economically.

\subsection{Identity and automated social categorization}
\begin{quote}
    "Let me note that identification is also a powerful factor in stratification; one of its most divisive and sharply differentiating dimensions. At one pole of the emergent global hierarchy are those who can compose and decompose their identities more or less at will, drawing from the uncommonly large, planet-wide pool of offers. At the other pole are crowded those whose access to identity choice has been barred, people who are given no say in deciding their preferences and who in the end are burdened with identities enforced and imposed \emph{by others}; identities which they themselves resent but are not allowed to shed and cannot manage to get rid of. Stereotyping, humiliating, dehumanizing, stigmatizing identities..." 

    - Zygmunt Bauman \cite{bauman2013identity}
\end{quote}

If we understand the social positionality of data annotation platform workers as being at the first end of this pole outlined by Bauman, a structural place characterized by a lack of choice in terms of work and social identity, their choices constrained by what work is available to cover basic material needs, and we understand certain people, e.g., elite technologists developing AI systems, in WEIRD societies as being able to choose among the large "pool of offers" of identities; then we can also see that the algorithmic and automated social categorization which happens in WEIRD societies: the stereotyping \cite{noble2018algorithms}, humiliation \cite{o2017weapons}, dehumanization \cite{buolamwini2018gender} and stigmatizing identities \cite{scheuerman2018safe} experienced by marginalized people is a similar phenomenon, except mediated and exaggerated by algorithms whose training data has been annotated by people occupying similar nodes in the structural hierarchy.

In other words, machine learning contributes to this cementing and crystallizing of social categorization that removes agency and choice from those whose position in the global hierarchy are subordinated. We draw from Foucault here, alluding to the pun in the title of our paper, who argues that questions of "truth" and knowledge cannot be separated from power, which is organized in specific ways in every society \cite{foucoult1975discipline, alcoff1996real}.  What is perhaps historically novel in the case of data annotation platform workers and data annotation is that through the operation of massive scale technology platforms that span the globe, these relationships of power - who gets to decide what the "correct" data label is, who develops and deploys AI systems - is truly globalized to a greater degree than in the past \cite{friedman2008modernities}. Platforms and machine learning enable this cross-global interaction of social identities in a novel way. We build on Miceli's call for broadening the field of responsible AI research from bias research towards an investigation of power differentials that shape data \cite{miceli2022studying}. 

Crucially, data annotation and machine learning, through the uncritical use of this system of automated social categorization, risks using measures and categories that are disconnected to how these concepts, categories and categorizations actually work \cite{monk2022inequality}. For example, race and gender categories in image datasets are presented as indisputable and self-evident for data annotation platform workers \cite{scheuerman2018safe, miceli2020between}. The requester of the annotations is granted the \emph{epistemic authority} to define these categories and how the data should be annotated (this epistemic authority is then further transferred to the outputs of the trained model) \cite{miceli2022studying}. The data annotation task instructions are typically designed and written by technologists in WEIRD countries, who for legal and policy reasons, rely on "state categories" of race and gender \cite{hanna2020towards}.

State categories are on the census, on identification cards, and on bureaucratic forms at hospitals, in prisons, in job applications and schools \cite{monk2022inequality}. These categories however are inadequate for capturing how social difference relates to inequality - which is one of our chief concerns here as it relates to explaining the social structural reasons for why some people are data annotation platform workers and some people are famous AI researchers. State categories play a crucial role in establishing who counts as what ethnoracial identity by defining these categories \cite{monk2022inequality} - and state categories form part of the \textit{habitus} of a society whether people agree with them or not. And, of course, data annotation workers are instructed to uncritically apply state categories to image datasets \cite{hanna2020towards, scheuerman2018safe}, text datasets \cite{kiritchenko2018examining} and other types of datasets that might require the annotation of social categories.

As Scott points out, all state categories are simplifications - like maps, which are designed to summarize precisely those aspects of a complex world that are of immediate interest to the map maker (or machine learning technologist) \cite{scott2020seeing}. Indeed machine learning as a field considers the machine learning model, its inputs, and the outputs typically in isolation, and abstracts away any context that surrounds the system \cite{selbst2019fairness}. The problem, and where the key issue of power must be accounted for, is when simplified state categories applied to data annotation, and thus algorithmic outputs, have the power to transform (have impacts on the real world) as well as serve their purpose as merely summarizing or simplifying the data which they have been given. This transformative power resides not in the machine learning model itself, but rather in the power possessed by those who deploy the perspective of that particular model \cite{scott2020seeing, miceli2022studying}. 

For example, as Scott continues, a private corporation aiming to maximize sustainable timber yields, profit, or production will map its world according to this logic and will use what power it has to ensure that the logic of its map prevails \cite{scott2020seeing}. In other words, states seek to turn the world its maps represent into actual physical changes that more closely resemble the logic of the maps. Likewise, the private technology platform industry seeking to maximize profit will use what power it has to ensure that logic of its machine learning models prevails.

The way human identity and social categorization work however are far more complicated than either state categories or machine learning models allow. As Lakoff argues, "Categorization is not a matter to be taken lightly. There is nothing more basic than categorization to our thought, perception, action, and speech \cite{lakoff2008women}." 


\section{This is why who annotators are is important (Individual Subjectivity's Influence on Social Categorization)}
The focus of our analysis here is the perception and judgment of social categories when data annotators are asked to identify when social categories apply to individuals or groups in data. A key consideration in how annotators-- and people more broadly-- categorize others socially is the sociocultural context of the individual doing the categorizing. Research in AI ethics and responsible AI has explored the influences of sociocultural difference on the design of AI systems, including on conceptions of what constitutes fairness itself \cite{sambasivan2021everyone}. In the context of data annotation, the interpretive lens that annotators apply has been investigated in relation to academic training \cite{sen2015turkers}, age \cite{diaz2018addressing}, global cultural region \cite{davani2023hate}, as well as other forms of social experience (e.g., \cite{waseem2016you, patton2019twitter}).

Importantly, a critical breakdown can occur in the labeling pipeline wherein errors rooted in mismatches between annotator interpretive lenses and the social context of source data are missed by ML developers, who typically operate according to a paradigm that treats annotators as interchangeable. One clear example of this is hate speech annotation of minoritized forms of speech, such as queer vernacular. Like many other sociolects, queer vernacular in the English-speaking world features reclaimed slurs in addition to unique or idiosyncratic phrases. As a result, situated, non-mainstream uses of language may be misjudged by annotators-- especially reclaimed language. One of example of this is work in natural language processing that focuses on both classifiers and data labels that skew more toxic for speech that resembles Black American English. Sap et al. investigated this phenomenon further, finding that, when annotators were given an explicit metric indicating that a data point resembles Black American English, biases completely disappeared \cite{sap2021annotators}. Of course there are at least two potential reasons for this. The first is that annotators may have exhibited a demand effect and adjusted their judgments so as not to appear racist. The second is that the metric may have helped annotators adjust their actual interpretation of the data.

Martin Sap's \cite{sap2021annotators} work raises the question of whether priming annotators with information about the author of data shifts the interpretive lens they use to apply judgements. Some research has begun to probe at overcoming this challenge. For example, D\'iaz and Amironesei point toward changes in annotation paradigms to account for minoritized use of language by disentangling annotator recognition of author 'voice' \cite{diaz2022accounting}. Although further research is required to understand the nature of the behavior Sap et al. observed, and more broadly determine the implications of cultural mismatch, it at least seems intuitive to expect a higher degree of misunderstanding or erroneously applied labels when annotators are presented with data that is far removed from their situated experience.

Beyond this, there is a broader question of the degree to which we can expect to match any source data to an annotator with the "correct" sociocultural lens. This is important because, even with carefully constructed task instructions and annotator training, we cannot transform annotation into a purely objective endeavor. Indeed, recent work in ML and NLP argues against any possibility of a completely objective form of annotation \cite{rottger2021two, basile2021toward}. \citet{basile2021toward} advocate for the adoption of a perspectivist approach to machine learning, which involves moving away from gold standard data and instead integrating the spectrum of human opinions and perspectives into knowledge representations. They also argue this for tasks that are otherwise considered to be objective, using the example of medical decision making (ibid). Still relative mismatches currently go largely ignored, in large part because data is taken out of context and presented to annotators with limited visibility into its origins.

\section{Socio-economic globalization (Labor Influence on Social Categorization)}

One reason the social facets of data annotation have only recently been explored lies in its scaled and globalized structure. Crowdsourced data annotation, as a practice fueling ML development, emerged as a blueprint for annotation with the creation of ImageNet \cite{deng2009imagenet}, which newly integrated human annotation through Amazon Mechanical Turk to supplement web-scraped data and enable quality checks \cite{denton2021genealogy}. The annotation process involved 49 thousand annotators and enabled a scale of work that would have been impossible through prior processes of hiring WEIRD undergraduate student labor \cite{gershgorn2017data}. Ultimately, the process was driven by a desire for scale and efficiency, which is reflected not only in the use of platforms and services for sourcing cheap labor, but also in the individual task components, which are designed as quick questions that can be responded to with simple clicks or responses.

The modular design and distribution of tasks relies on an implicit assumption of worker interchangeability, which others have pointed out is in tension with the use of crowdsourced data annotation for subjective tasks such as hate speech detection \cite{diaz2022crowdworksheets}. The result is a globally distributed workforce whose sociocultural variation remains largely unaccounted for. The issue isn't simply a matter of different cultural interpretations of meaning (which is itself a complicated challenge). Data annotation platform workers navigate economic and labor pressures by making best guesses as to the types of answers that data requesters want \cite{miceli2020between}. WEIRD dynamics take on a kind of Frankenstein form via a blend of 1) sociocultural biases, 2) differences in data annotation worker patterns of recognition, and 3) data annotation worker interpretations of what data requesters want as a force that mediates 1 and 2 \cite{miceli2020between, miceli2022data, denton2021whose, wang2022whose}. Another layer lies in the fact that gold data for training annotators and evaluating models often comes from the data requesters themseleves or in-house raters rather than outsourced ones. These processes are intended to be superficially agnostic to sociocultural perspectives embedded in source data and the sociocultural perspectives held by annotators.

\section{Annotation Task Design's Influence on Social Categorization}
Of the various influences on social categorization, annotation task design is the one that data requesters perhaps have the most control over. Röttger and Hovy point out two prevalent paradigms in annotation task design: one in which an annotator's individual subjectivity is maximally encouraged, and one in which individual subjectivity is maximally constrained \cite{rottger2021two}. In the context of hate speech detection, this could be represented by the distinction between asking annotators, "In your opinion, does this statement constitute hate speech?" or "Does this statement offend you?" compared with asking whether a statement contains a checklist of requester-defined features (e.g., insulting language, a named social group, etc.). In the first example, an annotator is asked to make a direct judgment. In the second, the annotator is intended to provide lower level information that the requester will use to deduce the presence of hatefulness, regardless whether the annotator personally believes the statement is hateful. The paradigm that maximally constrains subjectivity aims to produce more "objective" labels, however Röttger and Hovy critically point out that no degree of constrained design can completely eradicate vestiges of individual subjectivity \cite{rottger2021two}. It is equally important to note that subjectivity is not made objective through practices that seek to minimize inter-annotator disagreement (i.e. annotation guidelines) \cite{plank2014debatable}. In fact, one analysis showed that the majority of disagreements are due to legitimately hard cases \cite{plank2014debatable}. This further serves as evidence for the need to understand the influence of task design on social categorization.   

However, alternative approaches to this work should account of the subjectivity of the annotation task, which - at least in the context of text labels - is an interpretative task that can be cognitively demanding and requires perspective taking. Traditionally, annotations are done from the perspective of the annotator (interpreting the guidance from the data requester), without regard to the perspective or intention of the author of the original content. De-contextualized annotation work misses important nuances, e.g. friendly advocacy and moral justifications for a harmful cause \cite{friedman2021toward}, outrage related to injustice or sarcasm. A more constructive approach to data annotation is one that includes reason and nuance via qualitative data (e.g. in the form of interviews with teams of annotators that deliberate about conceptual understanding and definitions of the annotation task). Lost diversity of viewpoints often arises already with constraint definitions. For example, the label "empathy" can be interpreted as "care", "perspective taking" "feelings of compassion" etc. and is often directed at a specific target. However, that target can be 'victims of a war', but also 'supporters of a regime that disregards human rights'. Thus, understanding the intention of the author requires understanding the context of the author, which in turn includes culture, language and socio-economic variables. 


\section{Conclusion: The Future is not WEIRD but human}
We have outlined here the beginning of a genealogy and social theory of data annotation. This work seeks connections between the epistemic foundations of
1) machine learning, 2) the rise of computerized psychological studies and online platforms that can give psychology Big Data, 3) the problem of generalizing data from atypical college students in WEIRD countries, 4) the globalized market-dependence and dispossession of data annotation workers, 5) how structural inequality relates to social identity and "state" categorization, 6) how these are related to the social categorizations that data annotation platform workers are tasked with performing. We further draw connections between how this processes contributes to algorithmic social harms in WEIRD societies where algorithms are deployed by the state, corporations and other organizations in decision-making contexts. Such deployments can have severe consequences that also can contribute to stereotyping, stigmatization and identity formation. 

As de Oliveira and Baggs point out about psychology, the field as currently constituted is a fundamentally WEIRD enterprise and coming to terms with this is necessary if we wish to make psychology relevant for all humanity \cite{de2023psychology}. Likewise, if we wish to make machine learning or AI systems that are relevant, useful---or beneficial---for all humanity we must come to terms with the current exploitative practice of data annotation work. We must see data annotation workers as full human beings with unique qualities and annotation as deep work. They are not  distributions, variances to correct for, or categories that flatten their identities into numerically convenient measures of diversity. The pursuit of actual justice requires coming to terms with the global inequality which leads to the existence of vast numbers of data annotation workers to begin with. Research on data annotation has so far mostly involved a willingness to acknowledge and even revel in cultural difference without seriously challenging ongoing structural inequality \cite{benjamin2023race}. We argue that it is therefore imperative that further research be done in the area of structural inequality as it relates to data annotation practices. Last, we do not however endorse the automation of human annotations, as is currently being proposed as a solution by many as both a cost saving measure, and as a way to circumvent the epistemic issues outlined in this paper \cite{desmond2021semi}. Our argument is not a technical one. It is social and political.

\newpage
\bibliographystyle{ACM-Reference-Format}
\bibliography{disciplineandlabel}


\end{document}